\documentclass[letterpaper,10pt,journal]{IEEEtran}
\usepackage{blindtext}
\usepackage{graphicx}
\usepackage[bookmarks=false, colorlinks=false,
    linkcolor=black,
    filecolor=magenta,      
    urlcolor=cyan]{hyperref}   
\usepackage{amsmath}
\usepackage{amsfonts}
\usepackage{amssymb}
\usepackage[linesnumbered, ruled,vlined]{algorithm2e}
\SetCommentSty{mycommfont}

\newtheorem{definition}{Definition}

\author{Tauhidul Alam$^{1}$, Gregory Murad Reis$^{2}$, Leonardo Bobadilla$^{2}$, and Ryan N. Smith$^{3}$
\thanks{$^{1}$ T. Alam is currently affiliated with the State University of New York at Old Westbury. (e-mail:alamt@oldwestbury.edu)
}
\thanks{$^{2}$
 G. M. Reis and L. Bobadilla are with the School of Computing and Information Sciences, Florida International University, Miami, FL 33199, USA (e-mail:\{greis003, bobadilla\}@cs.fiu.edu)
}
\thanks{$^{3}$ R. N. Smith is with the Department of Physics and Engineering, Fort Lewis College, Durango, CO 81301, USA (e-mail:rnsmith@fortlewis.edu)}}
\begin{document}
%
\title{An Underactuated Vehicle Localization Method in Marine Environments}

\maketitle

\begin{abstract}

The underactuated vehicles are apposite for the long-term deployment and data collection in spatiotemporally varying marine environments. However, these vehicles need to estimate their positions (states) with intrinsic sensing in their long-term trajectories. In previous studies, autonomous underwater vehicles have commonly used vision and range sensors for autonomous state estimation. 
Inspired by the intrinsic sensing and the persistent deployment, we investigate the localization problem (state estimation) for an inexpensive and underactuated drifting vehicle called a {\em drifter}.
In this paper, we present a localization method for the  drifter making use of the observations of a proprioceptive sensor, i.e., compass.  We create the water flow pattern within a given region from ocean model predictions, develop a stochastic motion model, and analyze the persistent water flow behavior. Given a distribution of initial deployment states of the drifter at a particular depth of the water column  within the region and the water flow pattern, our method finds attractors and their transient groups at the given depth as the persistent behavior of the water flow. 
A most-likely localized trajectory of the drifter for a sequence of compass observations is generated based on the persistent behavior of the water flow and hidden Markov model. Our simulation results based on data from ocean model
predictions substantiate good performance of our proposed localization method with a low error rate of the state estimation in the long-term trajectory of the drifter.

\end{abstract}
\maketitle

\section{Introduction}
The application of underactuated underwater vehicles has attracted considerable interests in ocean monitoring~\cite{molchanov2015active}, ocean observation~\cite{smith2014controlling}, and coral reef surveying~\cite{xanthidis2016shallow,li2016data} since they require minimal resources in the actuation and sensing. Also, these vehicles are suitable for persistent deployment~\cite{alam2018deployment} in marine environments to tackle several oceanographic tasks. The fundamental requirement for autonomous underwater vehicles is the  ability to estimate their positions in a marine environment prior to addressing other tasks. This estimation of position in a marine environment for underwater vehicles is referred to as the localization problem. This localization problem is also  critical for the long-term autonomy of the underwater vehicles.
Therefore, we aim herein to solve the localization problem for an inexpensive and underactuated drifting vehicle called a {\em drifter} during its long-term deployment in a marine environment;
which is a challenging task in a GPS-denied and communication-challenged environment at the middle water column of an ocean. 



\begin{figure}
  \begin{center}
  \includegraphics[scale=0.0645]{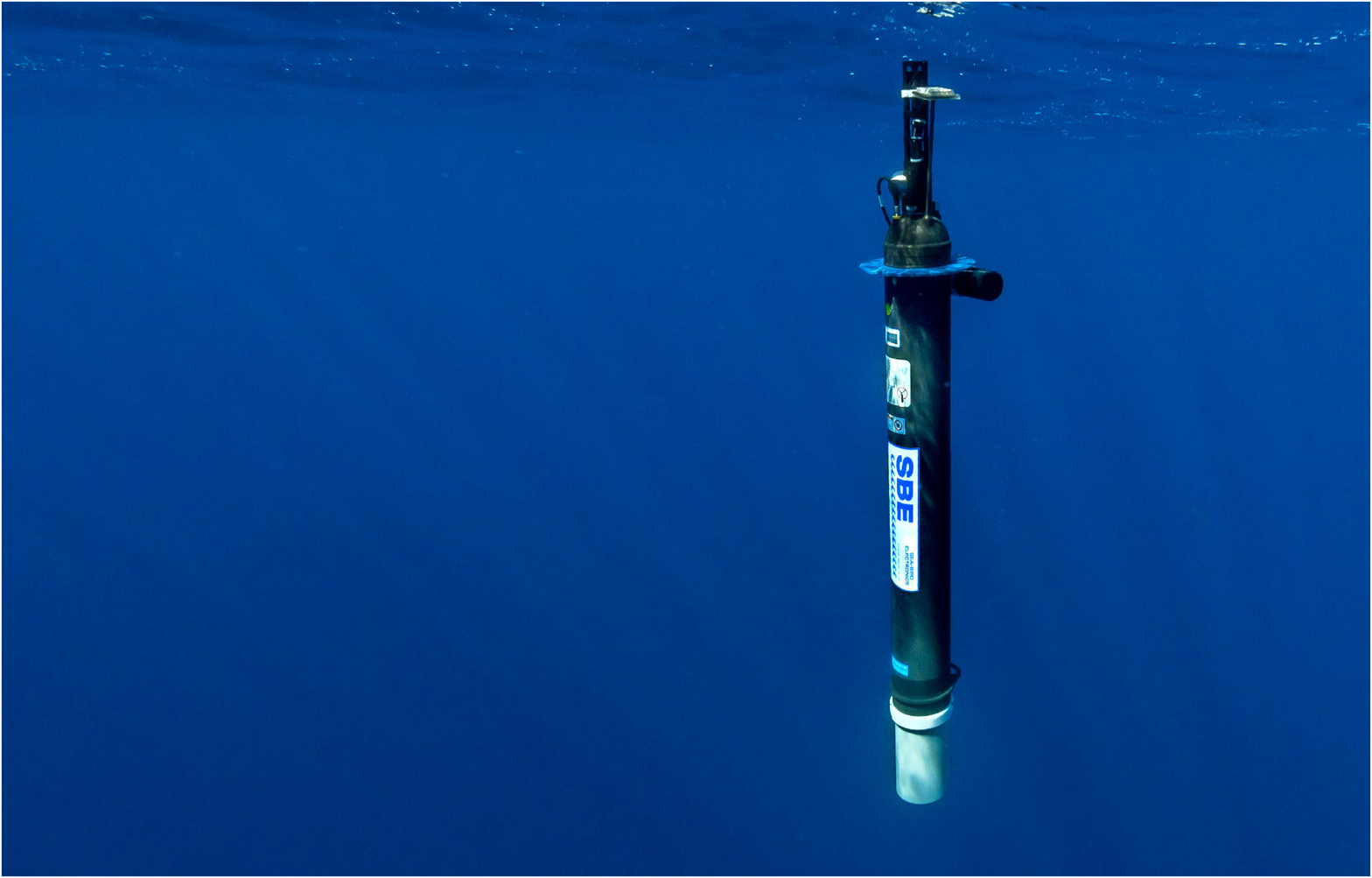}
    \hspace{2pt}
    \includegraphics[scale=0.38]{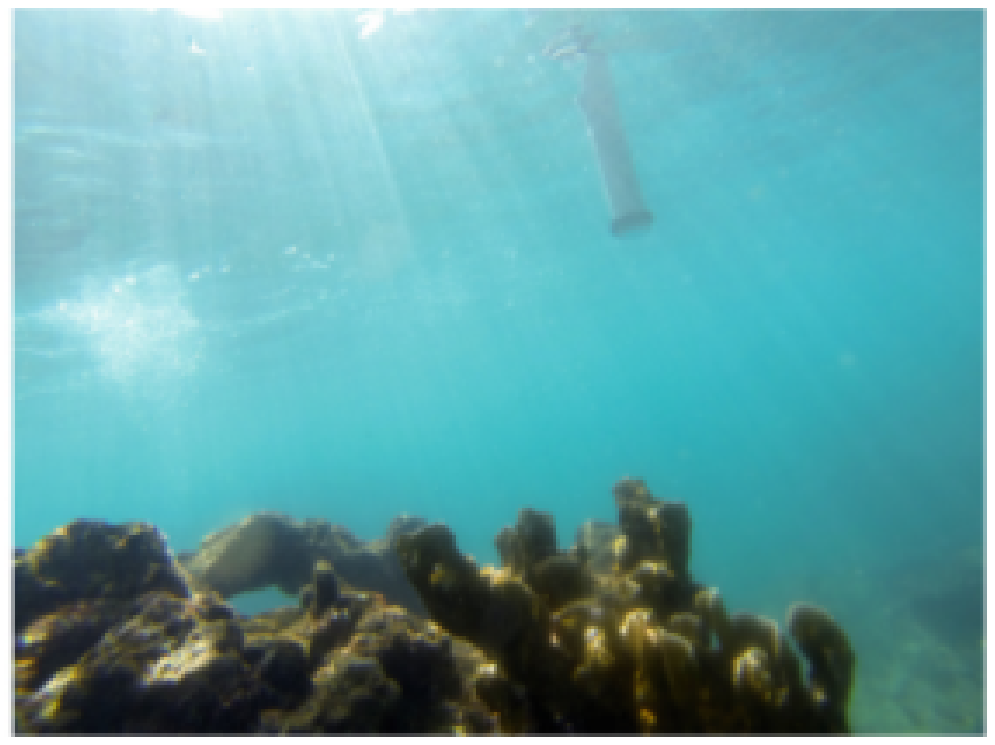}
  \end{center}
\caption{\label{fig:drifters} Two examples of drifters observing the marine environments at particular depths~\cite{pfloat,xanthidis2016shallow}.}
\vspace{2pt}
\end{figure}


The conventional underwater vehicles use data about a marine environment from extrinsic sensors such as camera, sonar, lidar and radar for determining their positions~\cite{palmer2009vision,maurelli2009sonar,bahr2009cooperative}. The use of these extrinsic sensors will result in excessive resource utilization and accompanying cost for solving the localization problem. Furthermore, these sensors will not work in various underwater scenarios such as murky water, fluctuating water temperature, and highly variable water currents. Moreover, the data transfer of these sensor-based localization methods suffers  from  low  bandwidth,  high  latency,  and  significant packet loss in the underwater environment.
These vehicles also make use of Doppler Velocity Logs (DVL) and GPS for precision localization near the water surface of a marine environment~\cite{corke2007experiments,karimi2013comparison}. However,  both GPS and DVL observations are unavailable after the mid-water column of the marine environment. Our minimally-actuated drifter can only rely on a proprioceptive sensor, e.g., compass, and water currents for finding its position while it floats at a specific depth of a marine environment.

Our localization problem for drifters involves the potential trajectory estimation for a sequence of sensor observations taking into account the long-term ocean dynamics. 
The drifter is equipped with proprioceptive sensors such as an inertial measurement unit (IMU) and compass, temperature sensor, WiFi communication module, and a Raspberry PI computing unit. 
The drifter floats on the water carried by currents, waves, and wind and collects data on environmental attributes such as temperature, salinity, turbidity, and chlorophyll contents.  
This drifter has an endurance of days to weeks during its deployment. These vehicles can control the buoyancy to change its depth for drifting at a particular water current layer. They are also called {\em profiling floats}~\cite{smith2014controlling}.
In this scenario, the localization is crucial since the drifter needs to know its locations along its long-term trajectory in a marine environment. The localization of the drifter is also challenging because of the spatiotemporal dynamics of the environment, the disturbances caused by ocean currents,  and the limited sensing and actuation capabilities of the vehicle itself. Two examples of drifting vehicles to observe marine environments are shown in Fig.~\ref{fig:drifters}.

The contribution of our paper is as follows. We propose a data-driven localization method for the underactuated drifter. Our proposed method creates a spatiotemporally varying water current field~\cite{medagoda2015autonomous}, develops a stochastic model, and finds the long-term water flow pattern.  Then, we generate a most likely trajectory of the drifter in a marine environment for localizing the drifter. A dynamical system technique called the {\em generalized cell-to-cell mapping} (GCM)~\cite{hsu2013cell} is applied for finding the long-term behavior of the water flow from the projected motion along with uncertainties in the form of a Markov Chain. 
Given a distribution of initial deployment locations of the drifter at a specific depth of the water column and a sequence of compass observations, our method finds attractors and their transient groups as the long-term behavior of the environment.  Then, our problem is framed as a Hidden Markov Model over the space of the environment and the compass observations. Finally, we
apply the {\em Viterbi algorithm}~\cite{forney1973viterbi} to find the most likely sequence of states of the drifter trajectory for the given sequence of its compass observations.


\section{Background}
\label{sec:rw}

Several AUV localization methods apply {\em recursive Bayesian filters}, e.g.,  particle filters~\cite{ko2012particle,maurelli2008particle} and Kalman filters~\cite{alcocer2007study,karimi2013comparison} utilizing   
 sensor data such as Doppler velocity log (DVL), inertial navigation system (INS), acoustic sensors, and range sensors. However, these methods are computationally expensive and require a large amount of memory capacity. Also, a minimally-actuated and resource-constrained underwater vehicle like a  drifter which is designed for long-term deployment~\cite{alam2018deployment}, cannot avail the requirements of these Bayesian filters based methods.
In~\cite{medagoda2015autonomous,medagoda2016mid}, the velocity analysis of spatiotemporally varying water current correlating with the velocity of the vehicle is incorporated to limit error growth in the position estimation. These closely related works are applicable for deep-diving and expensive AUVs that estimate the position through several sensor data fusion. Moreover, a number of closely related limited sensing localization methods for mobile robots have been proposed in~\cite{o2007localization,alam2018space,erickson2008probabilistic} where the robots entail less memory and computation to solve their localization task. The authors of this stream of research consider the motion model of simple ground robots whereas we take into account the dynamics of an underwater drifter combined with the water flow pattern.

A method to localize underwater
gliders in observation networks is proposed in~\cite{sun2016localization} utilizing the travel time of acoustic signals from near-surface acoustic sources.
An algorithm is developed to estimate an AUV's position
from the time difference of
arrival measurements in a long baseline acoustic
positioning system~\cite{thomson2017modeling}.
Range-based simultaneous AUV and multi-beacon localization in the presence of unknown ocean currents is presented in~\cite{bayat2016range}. A cooperative positioning algorithm is proposed for a fully mobile network of AUVs   that perform acoustic ranging and data exchange with one another in extended duration missions over large areas~\cite{bahr2009cooperative}.
The acoustic system methods discussed above need expensive infrastructure which includes the long base-line (LBL) and short or ultra short base-line (SBL) systems along with a set of transponders and modems, and the support of a surface ship.
This expensive infrastructure limits the application of the acoustic system methods in long-term missions.

Vision-based localization approaches of underwater robots are proposed in~\cite{palmer2009vision,carreras2003vision} which rely on the detection of a collection of features or patterns in subsea offshore structures for the pose estimation of robots. 
These vision-based underwater robot localization methods work well in experimental setups and structured environments. However, it is hard to deploy these tools in deep-sea or in unstructured environments. Map-based localization methods in structured environments using bio-inspired flow sensing are presented in~\cite{muhammad2017underwater,fuentes2017map,muhammad2015flow} that exploit the extraction of flow features and the speed or pressure estimation.
An IMU and a laser-based vision system is utilized for the localization of underwater vehicles~\cite{karras2007localization}. An experimental work  is described in~\cite{corke2007experiments} on the comparison of
acoustic localization with GPS for surface operation, and the comparison of acoustic and visual methods for underwater
operations. In our previous work~\cite{reis18oceans}, we present a tracking algorithm using a whitening technique for estimating the state trajectory of an AUV from the GPS observation history to infer its location. 

Our work is motivated by the analysis of mid-water current field aided localization methods~\cite{medagoda2015autonomous,medagoda2016mid}. Furthermore, this work is proposed for an underactuated  drifter with intrinsic sensing during its persistent deployment. 


\section{Model and Problem Definition}
\label{sec:pre}

In this section, we model the representation of the environment, and present the motion model
for the specific underactuated underwater vehicle. Then, we formally state the problem we consider.

\subsection{Model Definition}
We consider a 2-D environment where a workspace is a marine environment at a particular depth or current layer of the water column denoted as ${\cal W}=\mathbb{R}^2$. We also consider the stability of the vertical water current profile in the environment. Let ${\cal O}$ be the land and littoral region of the environment which is considered an inaccessible region for the drifter.  The free water space of the marine environment at a given layer is composed of all navigable locations for the drifter, and it is defined as $E = {\cal W}  \setminus {\cal O} $.
We discretize the workspace ${\cal W}$ as a 2-D grid. This grid is also called a cell workspace as
this discretized grid is a collection of cells.  
Each grid point has a geographic coordinate in the form of
longitude and latitude $(x_t, y_t)$ in which $x_t, y_t \in \mathbb{R}$. The geographic coordinate
of each grid point represents the center of an equal-sized cell $z$.  Hence, each cell in the grid is represented
as $z = (x_t , y_t)$.
We model each drifter as a point robot without considering its orientation. The state space of the drifter is
denoted by $X = E$. A state of the drifter in the state space is indexed by a cell index $z \in \{1,\ldots,N\}$ where $N$
represents the total number of cells in $X$. Let $Z = \{1,\ldots, N\}$ denote the set of all cells in the state space. Let $x_I$ be a known initial deployment state of the drifter. A probability distribution of initial deployment states of the drifter is defined as $\pi$. A state trajectory of the drifter is denoted as $\tilde{x} : [0, t]  \rightarrow X$ for a finite time interval $[0, t]$. Let $Y$ be the {\em observation space} of compass output values, which include eight directions (N,  NE,  E,  SE,  S,  SW,  W,  NW)  and  a  ninth  idle  operation (staying at the same location). An {\em observation history} of the compass for the drifter is defined as $\tilde{y} : [0, t]\rightarrow Y $.

\subsection{Problem Formulation}
We consider that the drifter moves autonomously in the free water space $E$ according to the currents or waves of the marine environment. In each cell $z$ of $E$, except the boundary ones, we consider the simplified scenario that a drifter has a total of nine actions, based on the currents, wind, and waves at the specific layer of the water column. From a non-boundary cell $z$, the set of actions for the drifter are moving towards the eight neighboring directions and the idle action. We assume that the drifter moves from one cell $z$ to another cell $z'$ in the free water space $E$ following one of the nine actions, considered as the steady motion of the drifter. We include noise and uncertainty in the movement along with the steady motion to account for the modeling error and unmodeled dynamics. When the drifter collides with a boundary cell $z$, then it will either stay in the same cell or move to one of the neighboring cells. All the potential options are assumed to have uniform probability.

The drifter makes use of a proprioceptive sensor named compass and its passive movements based on the long-term ocean dynamics for its state estimation.
We assume that the drifter keeps track of the observations of compass readings to estimate its state or position. The drifter motion is derived from the predicted ocean model. The long-term ocean current flow will help us understand the potential trajectory of the drifter associated with the history of compass observations.  In the long run, this trajectory will provide the estimation of the probable final state of the drifter in the mid-water column of a marine environment. In this context, we formulate our localization problem for the drifter as follows.


 \noindent {\bf Problem 1. Localization of a drifter using ocean dynamics and proprioceptive sensing:}

 \noindent {\em Given the long-term ocean dynamics  
  and the sensing observation history $\tilde{y}$ of compass readings for a drifter, reconstruct the most likely localized trajectory $\tilde{x}$ of the drifter.}

\section{Method}
\label{sec:meth}
In this section, we detail our method for solving the problem stated in Section~\ref{sec:pre}.

\subsection{Data Acquisition}
\begin{figure} [ht!]
  \begin{center}
     \includegraphics[width=0.25\textwidth]{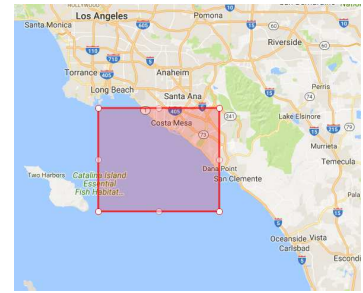}
  \end{center} 
 \caption{\label{fig:area} The area of interest in the SCB region, California.}
 \end{figure}

We use the Regional Ocean Modeling System (ROMS)~\cite{shchepetkin2005regional} predicted oceanic current data  in the Southern California Bight (SCB) region, California, USA, as illustrated in Fig.~\ref{fig:area}, which is contained within $33^\circ 17' 60''$ N to $33^\circ 42'$ N and $-117^\circ 42'$ E to $-118^\circ 15' 36''$ E. The model predictions of water current used herein are from July 2011.
ROMS is an open-source ocean model that is widely accepted and supported throughout the oceanographic and modeling communities. Furthermore, the model was developed to study ocean processes along the western U.S. coast which is our area of interest. The ROMS current velocity prediction data are provided at depths from $0$ m to $125$ m and for $24$
hours of the forecast for a consecutive number of days.
The four dimensions of the $4$-D ROMS current velocity prediction data consist of three spatial dimensions, e.g., longitude, latitude, and depth, associated with time. The three velocity components of
oceanic currents are the northing current ($u$), the easting current ($v$), and the vertical
current ($w$). These velocity components
are given based on the four dimensions (time, depth, longitude, and latitude). We assume that the vertical current velocity ($w$) is zero while a drifter floats at a particular current layer.
 We are using herein the water current velocity prediction data at a particular depth for a specific time.

\subsection{Create the Water Flow Pattern}

 We create a vector field at a particular depth or a given water layer in our marine environment of interest from the ROMS ocean current predictions data.  Ocean current velocity prediction data for a specific time and at a particular layer can be represented as a vector field. 
Let the vector field on a cell $z$ at a particular layer of the environment $E$  be $F(z)$.  
For a cell $z$ at a  particular layer, the easting velocity component along the latitude axis is denoted by $u(z)$, the northing velocity component along the longitude axis is denoted by $v(z)$, and the vertical velocity component is denoted by $w(z)$. 
The vector field based on two velocity components for a cell $z$ at a given layer  is specified as: 
\begin{equation}
F(z) = [u(z), v(z)]. 
\end{equation}
The vertical velocity component of the ocean current $w(z)$ at the corresponding layer is considered zero. Thus, we create the vector field for the given layer. Then, we find flow lines of the water flow from the vector field.
Flow lines of the water flow over the vector field $F$ are the trajectories or paths traveled by a point robot at the given layer whose velocity field is the vector field. 

To find the water flow pattern, we need all flow lines from cells in $Z$ at the given layer for a small time step $\Delta t$  so that we can map one cell to another cell based on these flow lines.  To calculate the next mapped cell $z(\Delta t)$ after a small time interval $\Delta t$ from each initial cell at time zero $z(0)$, we use the Euler integration method as follows:

\begin{equation}
z(\Delta t) = z(0)+ \Delta t \hspace{2pt} F(z(0)). 
\end{equation}

It gives the endpoint of the flow line from the initial cell $z$ after the small time $\Delta t$. After that, we use the Euclidean distance for locating the nearest cell from this endpoint. This nearest cell $z'$ becomes the next mapped cell of the initial cell $z$. Following this process, we obtain all flow lines for the small time step $\Delta t$ at the given layer. Finally, we get the next mapped cell for each cell in the environment $E$ at the corresponding layer.

\subsection{Long-Term Water Flow Analysis}


Given the cell mapping from the vector field and flow lines for the given 2-D current layer in the
previous step, we apply the GCM method~\cite{hsu2013cell,hong1999crises} for finding the long-term behavior of the created water flow pattern. Let $r$ be the probability of the steady or perfect motion of the water flow. Once we add the uncertainty in the water flow, we get a set of mapped cells at a particular layer for each cell $z$. Let $A(z) \subset Z$ represent the set of mapped cells at the given layer of a cell $z$ and $p_{zz'}$ denote the mapping probability of cell $z$ being mapped into one of the mapped cells $z'$.  The mapping probability $p_{zz'}$ has the following properties:

\begin{equation}
p_{zz'} \geq 0, \hspace{10pt} \sum_{z'\in A(z) } p_{zz'} =1.
\end{equation}

For a non-boundary cell $i$,  the mapping probability for the perfect motion $p_{ij} = r$ from cell $i$ to cell $j$ at the same layer, and the mapping probability for imperfect motion $p_{ij} = \frac{(1-r)}{(|A(i)|-1)}$ from cell $i$ to cell $j$ at the same layer. The calculated next mapped cell from the previous step identifies the mapped cell for the perfect motion. For a boundary cell $i$, we select all neighboring cells including the same boundary cell as a set of potential mapped cells with a uniform probability. Due to the nature of this cell mapping, the system evolution of GCM is expressed as:

\begin{equation}
p(n+1)= P p(n) \text{ or } p (n) = P^n p(0),
\end{equation}
\noindent
where $P$ is the one-step transition probability matrix, $P^n$ is the $n$-step transition probability matrix,  $p(0)$ is the initial  probability distribution vector, $p(n)$ is the $n$-step  probability distribution vector. Let $p_{ij}$ be the ($i,j$)-th element of 
$P$. It is called the one-step transition probability from
cell $i$ to cell $j$. Let $p_{ij}^n$ be the ($i, j$)-th element of $P^n$. It is called the $n$-step transition probability from cell
$i$ to cell $j$. If it is possible, through
the mapping, to go from cell $i$ to cell $j$, then we call cell $i$
leads to cell $j$, symbolically $i\Rightarrow j$. Analytically, cell $i$
leads to cell $j$ if and only if there exists a positive
integer $m$ such that $p^m_{ij}> 0$. 
If  cell $i$ leads
to cell $j$ and cell $j$ leads to cell $i$, then it is said that
cell $i$ communicates with cell $j$ or cell $j$ communicates
with cell $i$. This will be denoted by $i \Leftrightarrow j$.

This system evolution of GCM leads to a homogeneous finite Markov chain which determines the long-term behavior of the marine ecosystem.
To better understand the properties of GCM~\cite{hong1999crises}, we discuss some relevant definitions in the following:

\begin{definition}{(Persistent Cell)\quad} 
A cell $z$ is called a {\em persistent cell} if it has the property that when the system is in $z$ at a certain moment, it will return to $z$ at some time in the future.
\end{definition}
\begin{definition}{(Transient Cell)\quad} 
A cell that is not  persistent is called a {\em transient cell}. It leads to a persistent group in some number of steps.
\end{definition}

\begin{definition}{(Persistent Group or Attractor)\quad}
A set of cells that is closed under the mapping is said to form a { \em persistent group} if and only if every cell in that set communicates with every other cell. Each cell belonging to a persistent group is called a persistent cell. A persistent group is also termed as an attractor.
\end{definition}

Let $g$ be the total number of persistent groups in the system at a given layer. Let $B_i$ be an $i$-th persistent group or a set of persistent cells in $i$-th group where $B_i \subset Z$ and $i \in \{1,\ldots,g\}$. Thus, the set of all persistent groups or attractors at the given layer is denoted as ${\cal B}=\{B_1,\ldots,B_g\}$. If a transient cell $j$ leads to the $i$-th persistent group $B_i$, then we call $B_i$ a {\em domicile} of cell $j$.  A transient cell can also have several domiciles. 

\begin{definition}{(Single-domicile and Multiple-domicile)\quad}
Those transient cells that have
only one domicile are called {\em single-domicile} transient
cells, and those that have more than one are
called {\em multiple-domicile} transient cells.
\end{definition}

All the single-domicile transient cells having one particular persistent group as their common domicile form the {\em domain of attraction} of that persistent group at the same layer. A multiple-domicile transient cell having two or more persistent groups as its domicile is a cell in the {\em boundary region} between the domains of attraction of these persistent groups. Transient cells are further divided into transient groups according to 
the number of domiciles they have. Let $B(j)$ where $j=\{1,2,\ldots,g\}$ be the set of all single-domicile transient cells having $j$-th persistent group as its domicile. We call this $j$-th {\em single-domicile transient group}. It populates the domain of attraction of $j$-th persistent group. Let $B(i,j)$ where $i,j=\{1,2,\ldots,g\}$, and $i<j$,  be the set of all multiple-domicile transient cells having $i$-th and $j$-th persistent groups as their domiciles. We call this $(i,j)$-th {\em two-domicile transient group}. The region populated by this group is called the boundary regions of $i$-th and $j$-th domains of attractions. Hence, we  define the set of transient groups at the given layer as ${\cal T}$.


First, we find all persistent groups ${\cal B}$ and all transient groups ${\cal T}$ from  all the cells $Z$ in the environment $E$ at the given layer. 
Let $N_p$ be the total number of persistent cells and $N_t$ be the total number of transient cells at the same layer. Hence, the total number of cells $N$ at the same layer can be defined as $N=N_p+N_t$. 
Let ${\cal L}$ be the set of all persistent cells and ${\cal M}$ be the set of all transient cells at the same layer. Therefore, the set of all cells at the same layer can be specified as:
\begin{equation}
Z= {\cal L} \cup {\cal M}.
\end{equation}



We associate herein between the system evolution of GCM and the directed graph theory.  To obtain the properties of GCM from this association,  Algorithm~\ref{alg:TE} takes as input a geometric description of the environment $E$, and the vector field $F$, the observation space $Y$, a sequence of compass observations $\tilde{y}$,  and a distribution of initial deployment states $\pi$. It returns the set of persistent groups  ${\cal B}$, the set of transient groups ${\cal T}$, and the most likely trajectory of the drifter $\tilde{x}$ for the observation history $\tilde{y}$ at the given layer of the environment. For the given layer of the environment, Algorithm~\ref{alg:TE} creates a directed graph $G$ without adding weights to the edges of $G$ from the set of cells $Z$. 
Additionally, for each cell $z \in Z$, it finds the geographic location ($x,y$) (line 4).  From this geographic location ($x,y$), it gets the location ($x',y'$) of the next mapped cell based on the vector field $F$ as explained before (line 5). Taking the motion uncertainty into consideration, it computes the set of mapped cells $Z'$ where $Z' \subset Z$ (line 6). All cells $z, Z'$ are added to the vertices set and their ordered pairs ($z,z'$), where $z' \in Z'$, are added the edges set of $G$ (line 7$-$8).

\begin{algorithm}
\caption{\label {alg:TE}\textsc{TrajectoryEstimation}($E,F, Y, \tilde{y}, \pi$)}
\DontPrintSemicolon
\KwIn{$E$, $F$, $Y$, $\tilde{y}$, $\pi$ -- Environment, Vector field, Observation Space, Observation history, Initial  deployment distribution }
\KwOut{${\cal B}$, ${\cal T}$, $\tilde{x}$ -- Set of persistent groups, Set of transient groups, Most likely trajectory}
$G.V \gets \emptyset, \quad G.E \gets \emptyset, \quad {\cal B} \gets \emptyset, \quad {\cal T} \gets \emptyset, \quad {R} \gets \emptyset$\;
\For{$i \gets 1 \text{ to } N$}{
$z\gets i$ \;
$x, y \gets \textsc{CellLocation}(z)$\;
$x', y' \gets \textsc{MappedCell}(x, y, F)$\;
$Z' \gets \textsc{MappedCellSet}(x', y')$ \tcp*[f]{Add uncertainty}\;
$G.V\gets G.V \cup Z' \cup \{z\}$\;
$G.E\gets G.E \cup \{(z,z')\mid z'\in Z'\}$\;
}
$S \gets \textsc{StronglyConnectedComponent}(G)$\;
$C \gets \textsc{TransitiveClosure}(G)$\;

${\cal B} \gets \textsc{FindPersistentGroups} (S,C)$\;
${\cal L} \gets \textsc{FindUnion} (\cal B)$\;
$\mathcal{M} \gets Z \setminus \mathcal{L}$\;
${\cal T} \gets \textsc{FindTransientGroups}({\cal B}, {\cal M}, C)$\;
$ P \gets \textsc{TransitionMatrix} (S,C,r)$\;
$ Q \gets \textsc{EmissionMatrix} (S,C,Y,r)$\;
$\lambda  \gets \textsc{HMM}(P, Q, \pi)$\;
$\tilde{x} \gets \textsc{Viterbi}(\lambda, \tilde{y})$\;
\Return{${\cal B}$, ${\cal T}$, $\tilde{x}$ }\;

\end{algorithm}

Next, it finds the set of strongly connected components $S$ from  $G$ using Tarjan's strongly connected component algorithm~\cite{tarjan} (line 9). We define the connectivity matrix as $C$ and it is calculated from the transitive closure of $G$ (line 10). From  $S$ and $C$, Algorithm~\ref{alg:TE} finds the set of $g$ persistent groups  ${\cal B}$ using the function \textsc{FindPersistentGroups} (line 11). In this function, if each vertex in a strongly connected component communicates to all other vertices in the strongly connected component then this strongly connected component is found as a persistent group and each cell of this persistent group is classified as a persistent cell. Otherwise, each cell of this strongly connected component is classified as a transient cell. 
Taking the union of $g$ persistent group sets, we get the set of all persistent cells ${\cal L}$ (line 12). Aside from all the persistent cells, the remaining cells from $Z$ represent the set of transient cells ${\cal M}$ (line 13). Thus, it classifies all the cells $Z$ in the environment $E$ at the given layer into the set of persistent cells ${\cal L}$ and the set of transient cells ${\cal M}$. To determine the set of single-domicile and multiple-domicile transient groups ${\cal T}$ using the function \textsc{FindTransientGroups} (line 14), we check if there is any path or connectivity from each transient cell to cells in all $g$ persistent groups according to the connectivity matrix $C$ of graph $G$.   
Accordingly, we find the set of transient groups ${\cal T}$. Finally, the set of persistent groups or attractors ${\cal B}$ and the set of transient groups or domains of attractions ${\cal T}$ characterize the long-term water flow of the marine environment at the given layer.   




\subsection{Generate the Most Likely Localized Trajectory for a Drifter}

Given the collection of persistent groups or attractors and their associated transient groups or domains of attractions as the long-term water flow of the marine environment at the given layer, we generate the most likely trajectory of the drifter $\tilde{x}$ for its localization in the environment. Generation of the most likely trajectory of the drifter for its state estimation is modeled as a hidden Markov model (HMM). This hidden Markov model is defined by six elements. These elements are the state space $X$, the observation space $Y$, the one-step transition probability matrix $P$, the emission transition or observation transition probability matrix $Q$, an initial state distribution $\pi$, and a compass observation history $\tilde{y}$. To characterize this model,  Algorithm~\ref{alg:TE} calculates the transition probability matrix $P$ from the strongly connected components $S$, the connectivity matrix $C$, and the probability of reliable or perfect motion $r$ (line 15). It also computes the emission transition probability matrix $Q$ from the elements mentioned before along with the observation space $Y$ (line 16). We obtain a compass observation history of the drifter $\tilde{y}$ simulating the Markov chain $P$ from a distribution of initial deployment states $\pi$.  An HMM object is characterized as $\lambda = (P, Q, \pi)$ (line 17). Given a sequence of compass observations $\tilde{y}$ and the HMM object $\lambda$, Algorithm~\ref{alg:TE} computes the optimal hidden state sequence $\tilde{x}$ that best explains the compass observation sequence $\tilde{y}$ (line 18). This optimal hidden state sequence or the most likely trajectory of the drifter is computed using the Viterbi algorithm~\cite{forney1973viterbi}, which is a dynamic programming algorithm. This most likely hidden state sequence is also termed as the {\em Viterbi path}. This trajectory of the drifter represents its localized trajectory for the given compass observation sequence. Thus, we generate the most likely localized trajectory of the drifter from deterministic (a known initial state $x_I$) and probabilistic (a set of initial states including the neighbors of the original initial state) distribution  of initial deployment states $\pi$ for different compass observation sequences.

\section{Simulation Results}
\label{sec:res}
\begin{figure}[ht!]
\begin{center}
\begin{tabular}{cc}
\hspace*{-8pt}\includegraphics[scale=0.3]{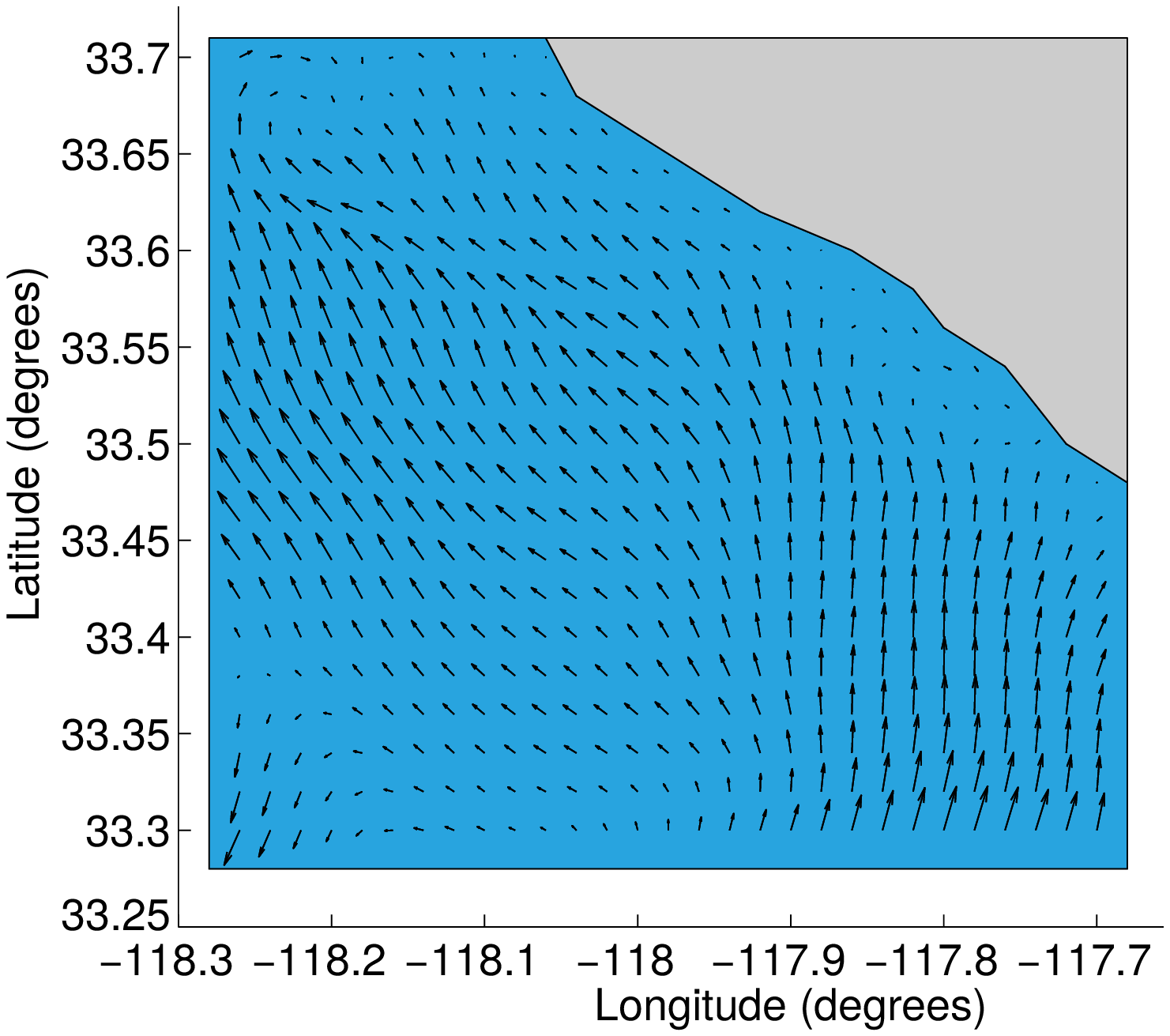} &
\hspace*{-10pt}\includegraphics[scale=0.3]{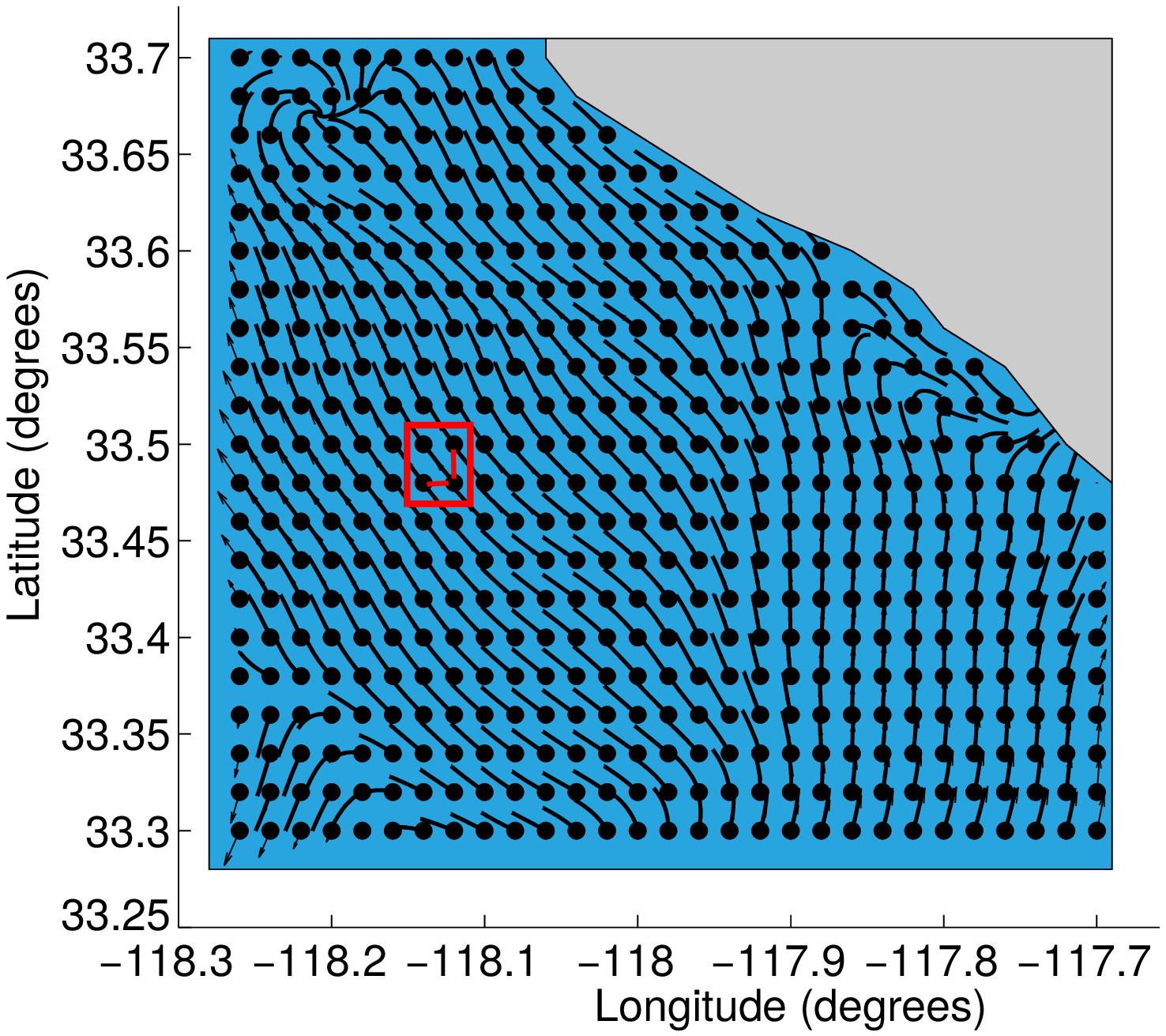}\\

\hspace*{-8pt}(a)  & \hspace*{-10pt}(b) \\

\end{tabular}
 \end{center}
\caption{\label{fig:vectorflowres} Vector field and flow lines: (a) The vector field generated from ROMS current prediction data; (b)  Flow lines for a small time step $\Delta t$ from all locations along with added red uncertain movements depicted inside the red box.}

 \end{figure}

We validated our method through simulations using the ROMS~\cite{shchepetkin2005regional} ocean current predictions data in the SCB region. An ocean environment was considered as the simulation environment for the drifter movements having one $2$D water current predictions of a given layer (i.e., $10$ m).  The  $2$D ocean water space was tessellated into a grid map. The resolution of the grid map was set to $21\times29$. The vector field from the ocean current predictions was calculated at the given layer of the environment. The flow lines of the ocean current data were generated through the Euler numerical integration method from all locations of the water space at the given layer for a small time $\Delta t$ which was the time to pass a cell. Thus, we found the cell mapping of each cell location of the water space at the corresponding layer. The vector field and the flow lines for this vector field of the simulated environment are illustrated in Fig.~\ref{fig:vectorflowres}.

\vspace*{5pt}

\begin{figure}[ht!]
\begin{center}
\begin{tabular}{cc}
\hspace*{-8pt}\includegraphics[scale=0.3]{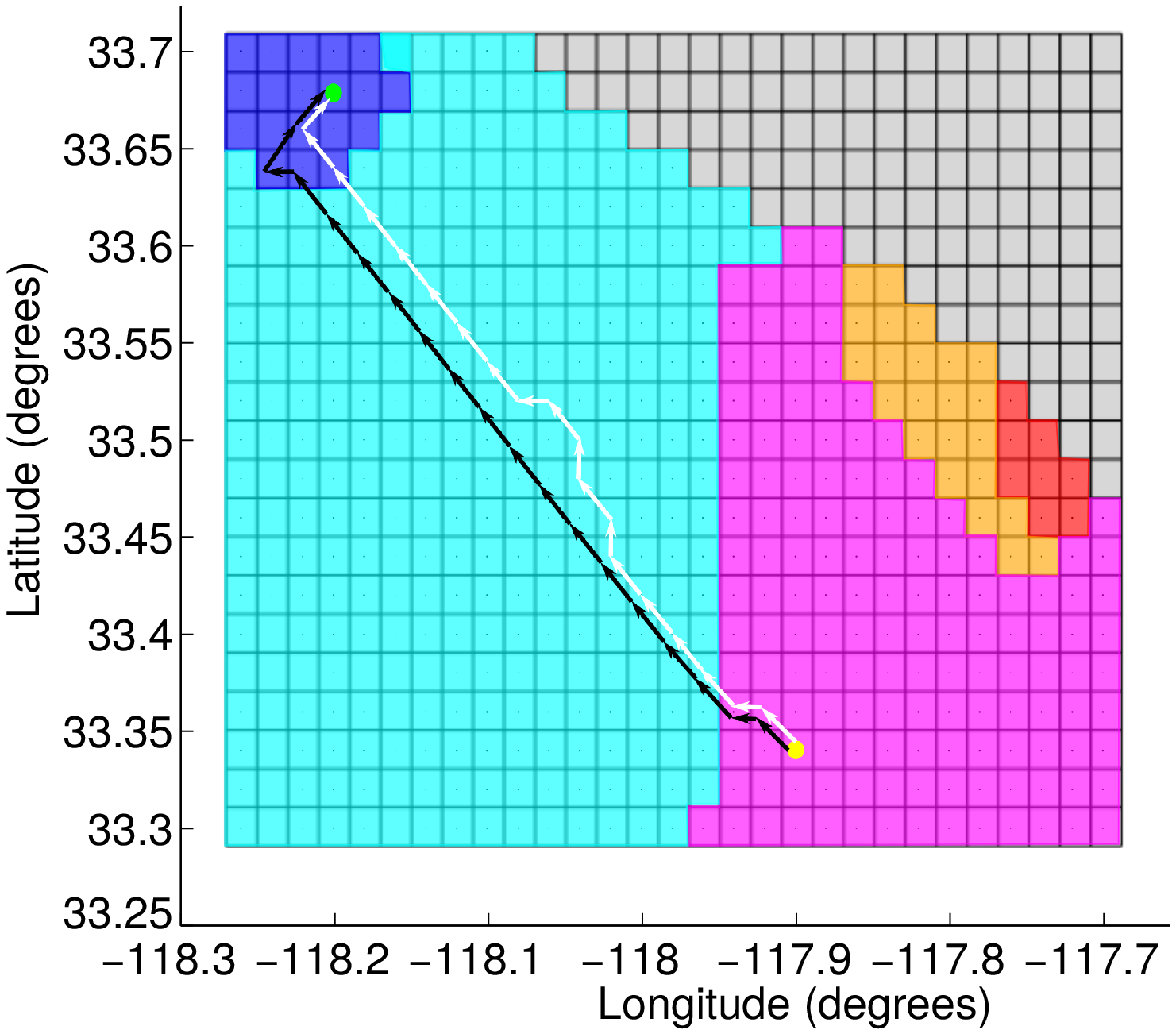} &
\hspace*{-10pt}\includegraphics[scale=0.3]{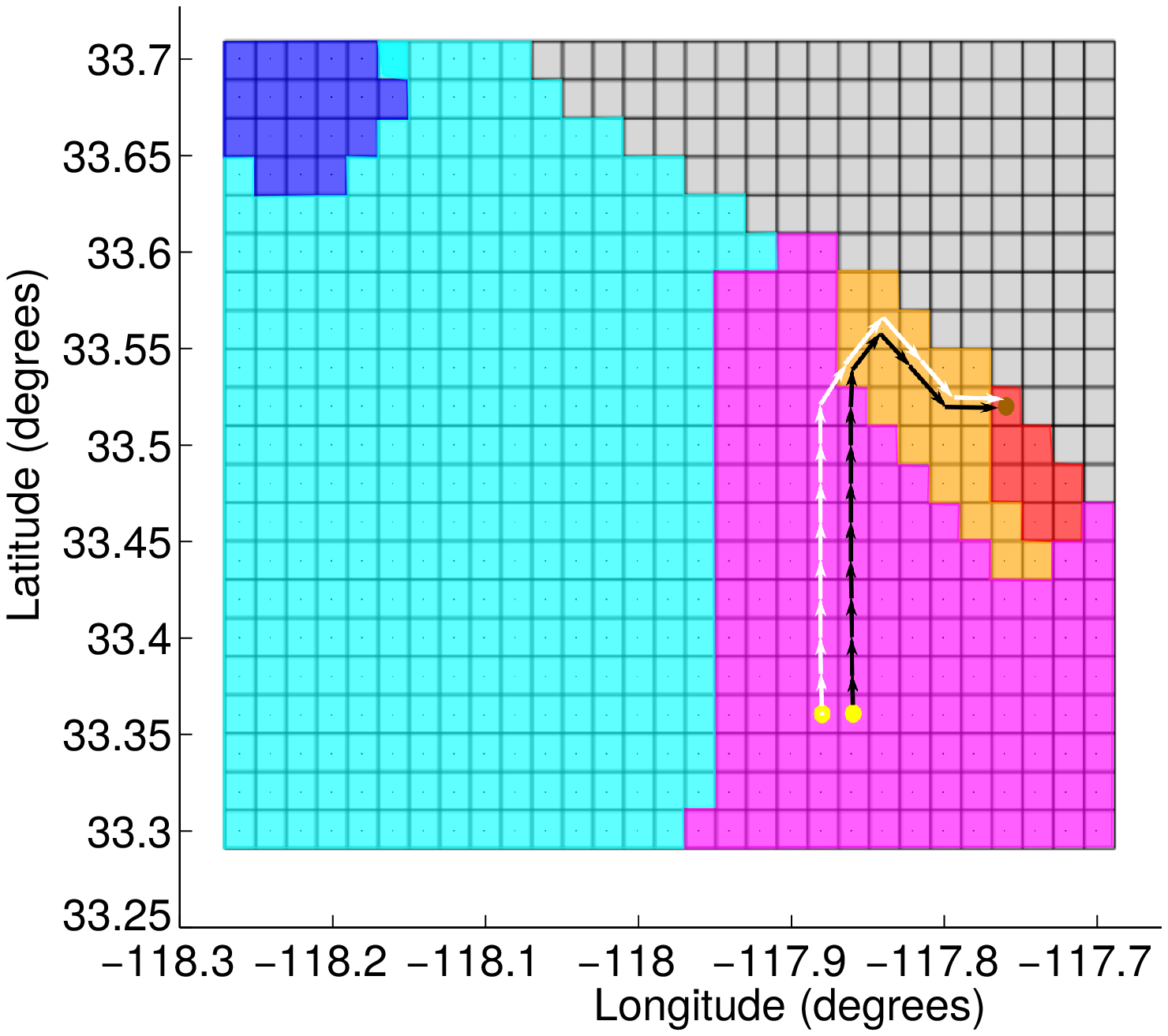}\\

\hspace*{-8pt}(a)  & \hspace*{-10pt}(b) \\

\end{tabular}
 \end{center}
\caption{\label{fig:localizationres} (a)$-$(b) Attractors (blue and red regions) and associated transient groups (the cyan region for the blue attractor, the orange region for the red attractor,  and the magenta region for both attractors); (a) The white original trajectory and the black most likely trajectory for the given compass observation sequence/history and  the deterministic distribution of the yellow initial location; (b) The white original trajectory and the black most likely trajectory for the given compass observation sequence/history and the probabilistic distribution of the yellow initial locations.  Both trajectories localized at the green final locations.}

 \end{figure}
 
 \vspace*{5pt}
 We implemented Algorithm~\ref{alg:TE} in a simulation to find the long-term water flow behavior of the considered water current layer and generate the most likely trajectory of the drifter based on the long-term water flow in the simulated environment. From the simulation run, we found two persistent groups and three transient groups where two of them are single-domicile transient groups and one is a multiple-domicile transient group.  The long-term behavior of the simulated environment and two instances (both deterministic, i.e., a known initial state and probabilistic, i.e., the nondeterministic neighboring states around the initial deployment state) of the generated most likely localized trajectory of the drifter are shown in Fig.~\ref{fig:localizationres}. 


\begin{figure} [ht!]
\begin{center}
\hspace{-10pt}\includegraphics[scale=0.28]{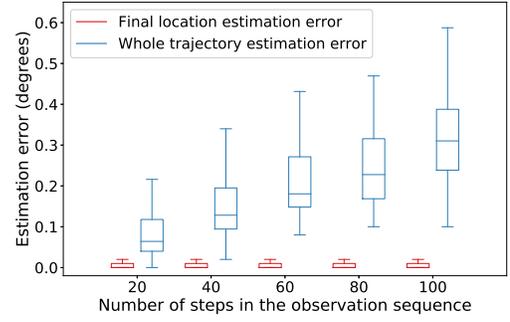}
\end{center}
\caption{\label{fig:dler} Comparison of the estimation error in the final location and the whole trajectory for the various number of steps in the compass observation sequence.}
\end{figure}

\begin{figure}[ht!]
\begin{center}
\hspace{-20pt}\includegraphics[scale=0.28]{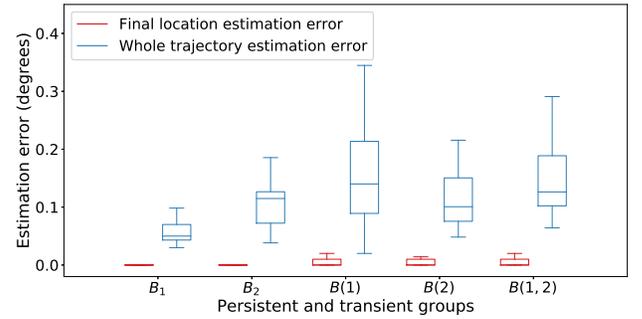}
\end{center}
\caption{\label{fig:drer} Comparison of the estimation error in the final location and the whole trajectory for different long-term regions (persistent groups or attractors and transient groups) of the environment.}
\end{figure}

\begin{figure}
\begin{center}
\hspace{-20pt}\includegraphics[scale=0.28]{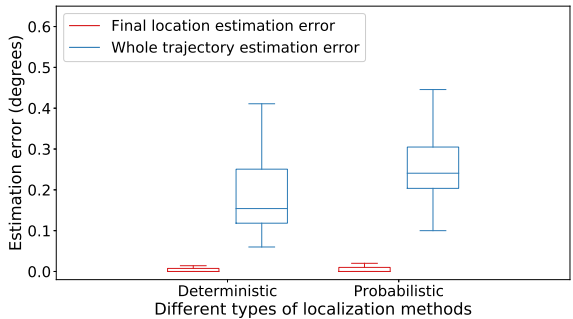}
\end{center}

\caption{\label{fig:dpler} Comparison of the  estimation error in the final location and the whole trajectory for deterministic and probabilistic localization methods.}
\end{figure}

For the quantitative analysis of our localization method, we calculated the Euclidean distance between the generated and actual final locations of the drifter in its trajectory as the final location estimation error. For the whole drifter trajectory, we also calculated the Euclidean distance between the generated and actual locations for each observation and summed up these distances for all observations that represents the whole trajectory estimation error. 
This is also the total deviation of the generated trajectory from the original trajectory of the drifter. 
For each of the five number of steps ($20$, $40$, $60$, $80$, $100$) in the compass observation sequence, we ran the simulation $50$ times using the deterministic method  and recorded the estimation error.
Then, we compared the results of estimation error for the final location of the trajectory and for the whole trajectory with respect to these various number of steps ($20$, $40$, $60$, $80$, $100$) in the compass observation sequence which is illustrated in Fig.~\ref{fig:dler}. 
We computed the same error for different long-term regions (attractors $B_1$, $B_2$, and transient groups $B(1)$, $B(2)$, $B(1,2)$) of the environment for $40$ steps in the compass observation sequence from $20$ deterministic simulation runs and show their comparison in Fig.~\ref{fig:drer}.
We also ran the simulation $50$ times for $50$ steps in the compass observation sequence using both deterministic and probabilistic methods in terms of the initial states and stored the estimation error for the final location of the trajectory and for the whole trajectory.  The results of the estimation error of deterministic and probabilistic localization methods  were compared  and are shown in Fig.~\ref{fig:dpler}.




\section{Conclusion and Future Work}
\label{sec:conc}
In this paper, we presented a localization method for an underactuated drifter floating at a particular depth of the water column in a marine environment.
First, we created the vector field to study the water flow pattern from the ROMS ocean current predictions data of the mid-water column. The water flow pattern was characterized as a stochastic model. The generalized cell-to-cell mapping method was applied to determine the persistent behavior of the water flow taking the stochastic model into account.
Based on the persistent behavior of water flow, we generated the most likely localized trajectory of the drifter for a  distribution of initial deployment states and the observation history from the compass inside the drifter using a hidden Markov model. Our results demonstrate that the final state estimation error is very low and the whole trajectory deviation increases with the increase of the observation length. Therefore, our localization method works considerably well although it depends on the observations from the intrinsic sensor of the drifter. 

In the future, we can do the temporal analysis of our method to evaluate the variability of the generated localized trajectory of the drifter over space and time. The neighboring current layers of a given layer in a marine environment can be incorporated for taking the vertical motion of our passive drifter into account in our localization method.  
 Our ideas can be also extended for the vehicles that have the capability of horizontal motion control in addition to the vertical motion capability. An example of such propeller-driven, long-range AUV is called {\em Tethys} which can be used to conduct  oceanic  missions  
over periods of weeks or even months without a ship~\cite{hobson2012tethys}. Moreover, our proposed  method is a solution to the {\em passive} localization problem. The inclusion of more motion capabilities to the profiling drifter in order to guarantee more accurate localization which will help develop a solution to the {\em active} localization problem.  

Our previous work~\cite{alam2018data} develops a control policy using a Markov decision process (MDP) on the fully observable states of the predicted ocean current model. We can combine this localization method with the previous control method by introducing a partially observable Markov decision process (POMDP) framework.




\section*{Acknowledgments}
We would like to acknowledge the financial support of a Florida International University Graduate School Dissertation Year Fellowship. We offer our thanks to the Coordination for the Improvement of Higher Education Personnel (CAPES), Brazil, for the doctoral scholarship. This work is also supported in part by the U.S. Department of Homeland Security under Grant Award Number 2017-ST-062000002, by the Office of Naval Research Award Number N000141612634, and by the National Science Foundation MRI Award Number 1531322.

\bibliographystyle{ieeetr} 
{
\bibliography{oceans} 
}

\end{document}